\documentclass[lettersize,journal]{IEEEtran}
\newtheorem{Thm}{Theorem}[section]

\newtheorem{Def}[Thm]{Definition}

\newtheorem{Rem}[Thm]{Remark}

\newcommand{\R}{\mathbb{R}}
\newcommand{\Z}{\mathbb{Z}}

\newcommand{\set}[1]{\mathbf{#1}}
\newcommand{\opr}[1]{\mathcal{#1}}

\newcommand{\EZ}[1][]{e}
\newcommand{\fEZ}[1][]{e}

\newcommand{\widebar}[1]{\overline{#1}}

\newcommand{\exampleset}{\set{S}}

\usepackage{graphicx}   
\usepackage{epsfig} % for postscript graphics files
\usepackage[T1]{fontenc} % Gives an error without this package
\usepackage{amssymb} % For set notations like Natural numbers, whole numbers etc.
\usepackage{caption}
\usepackage{subcaption}
\usepackage{booktabs}
\usepackage{amsmath} % For symbols over equal to sign
\usepackage[pdf]{pstricks}
\usepackage{algorithm}
\usepackage[noend]{algpseudocode}
\usepackage{graphicx}
\usepackage{lipsum}%
\usepackage{xcolor}
\usepackage{hyperref}
\usepackage{float}
\usepackage{enumitem}
\usepackage[backend=bibtex,style=ieee,sorting=none,doi=false,isbn=false,url=false,eprint=false]{biblatex}
\newcommand{\RNum}[1]{\uppercase\expandafter{\romannumeral #1\relax}}

\usepackage{color}

\usepackage{mathtools}

\title{\LARGE \bf Overcoming the Fear of the Dark: Occlusion-Aware Model-Predictive Planning for Automated Vehicles Using Risk Fields}
\author{Chris van der Ploeg$^{1,2}$, Truls Nyberg$^{4,5}$, José Manuel Gaspar Sánchez$^{4}$, Emilia Silvas$^{1,3}$, Nathan van de Wouw$^{2}$
\thanks{$^{1}$Netherlands Organisation for Applied Scientific Research, Integrated Vehicle Safety Group, 5700 AT Helmond, The Netherlands. (${\tt \{chris.vanderploeg, emilia.silvas\}@tno.nl}$)}
\thanks{$^{2}$Eindhoven University of Technology, Dynamics and Control Group, Mechanical Engineering Dept., P.O. Box 513, 5600 MB, Eindhoven, The Netherlands.(${\tt \{c.j.v.d.ploeg, n.v.d.wouw\}@tue.nl}$)}
\thanks{$^{3}$Eindhoven University of Technology, Control Systems Technology Group, Mechanical Engineering Dept., P.O. Box 513, 5600 MB, Eindhoven, The Netherlands.(${\tt e.silvas@tue.nl}$)}
\thanks{$^{4}$KTH Royal Institute of Technology, Division of Robotics, Perception and Learning, School of Electrical Engineering and Computer Science, 100 44 Stockholm, Sweden.(${\tt \{trulsny, jmgs\}@kth.se}$)}
\thanks{$^{5}$Scania CV AB, 151 87 Södertälje, Sweden.} %($\tt truls.nyberg@scania.com$)}
\thanks{This work was partially supported by the Wallenberg AI, Autonomous Systems and Software Program (WASP) funded by the Knut and Alice Wallenberg Foundation.}
\thanks{This research has been carried out as part of the TECoSA Vinnova Competence Center for Trustworthy Edge Computing Systems and Applications at KTH Royal Institute of Technology.}
}
\date{March 2023}
\begin{document}
\maketitle
\thispagestyle{empty}
\pagestyle{empty}
\begin{abstract} 
As vehicle automation advances, motion planning algorithms face escalating challenges in achieving safe and efficient navigation. Existing Advanced Driver Assistance Systems (ADAS) primarily focus on basic tasks, leaving unexpected scenarios for human intervention, which can be error-prone. 
Motion planning approaches for higher levels of automation in the state-of-the-art are primarily oriented toward the use of risk- or anti-collision constraints, using over-approximates of the shapes and sizes of other road users to prevent collisions. These methods however suffer from conservative behavior and the risk of infeasibility in high-risk initial conditions.
In contrast, our work introduces a novel multi-objective trajectory generation approach. We propose an innovative method for constructing risk fields that accommodates diverse entity shapes and sizes, which allows us to also account for the presence of potentially occluded objects. This methodology is integrated into an occlusion-aware trajectory generator, enabling dynamic and safe maneuvering through intricate environments while anticipating (potentially hidden) road users and traveling along the infrastructure toward a specific goal. Through theoretical underpinnings and simulations, we validate the effectiveness of our approach. This paper bridges crucial gaps in motion planning for automated vehicles, offering a pathway toward safer and more adaptable autonomous navigation in complex urban contexts.
\end{abstract}
\begin{IEEEkeywords}
    Motion planning, situational awareness, occlusion awareness, model predictive control, artificial potential fields.
\end{IEEEkeywords}
\section{Introduction}~\label{sec:introduction}
As vehicles become more automated, motion planning algorithms face tougher demands. In ADAS systems currently on the market, the primary goal of motion planning is to maintain a safe distance from the vehicle in front, while adhering to a set speed limit and staying in the center of the lane. Any unexpected situation must still be identified and handled by the human driver. It is in human nature to be prone to errors due to, e.g., distraction or exhaustion~\cite{dingus_driver_2016}, causing safety hazards. As such, higher levels of automation are demanded to mitigate or prevent these human errors and to achieve higher safety levels.

When it comes to higher levels of automation, it is important for motion planning algorithms to exercise caution and carefully consider any potential risks. Risk, defined by industry standards such as those by the automotive industry~\cite{1400-1700_iso_2018,1400-1700_iso_nodate}, represents the probability of harm from hazardous events. Such notion of risk is influenced by both the severity of the event and the likelihood of it happening. One of the risks most studied in the field of motion planning is the risk of collision with observable road users. For example, the works in~\cite{tolksdorf_risk_2023,khonji_risk-aware_2020,nyberg_risk-aware_2021} employ the risk of collision (based on, e.g., a time-to-collision (TTC) or a time-to-react (TTR) criterion) to timely replan actions to fully avoid, minimize or constrain this risk. The work in~\cite{geisslinger_ethical_2023} brings this notion of risk a step further, looking from the perspective of collective risk, i.e., ensuring that risk is properly constrained and distributed from an entire traffic system standpoint rather than the single vehicle standpoint. It is understandable to use constraints to prevent collisions or bound the risks, but it may not always be feasible to implement or adhere to them~\cite{brudigam_minimization_2021}.

In~\cite{van_der_ploeg_connecting_2023}, the notion of risk is extended to, not only incorporate a risk of collision, but augment this risk with situational comprehension, incorporating elements such as object classification, road layout, social norms, and traffic rules that may apply in certain conditions. The risk is then mapped to a risk field (formulated through the classic notion of artificial potential fields~\cite{ji_path_2017}) to be incorporated in the cost (to be minimized) of an optimal planning program, rather than incorporating it as a constraint. The work in~\cite{jensen_mathematical_2022} shows that these risk fields correspond to the risk perceived by a human driver to make decisions during driving. Therefore, when a collision is unavoidable, a minimal-risk maneuver can be selected, instead. Furthermore, the work incorporates the risk versus reward model (further elaborated in~\cite{van_der_ploeg_long_2022}) in a model-predictive program, allowing a quantitative multi-objective trade-off for risk planning which, by design, is also motivated to progress along a path. Similar to the formulation of constraints, however, most of these approaches require over-approximations of the considered entities (e.g., dynamic objects, road markings, sidewalks), which might make the motion planning problem more conservative.

Another vital component that contributes to risk-averse driving is the anticipation of unseen or occluded objects. Road collisions often occur due to the geometric features of the road, such as buildings and landscapes, obstructing the driver's view~\cite{zhang_safe_2021}. There are several factors that can lead to late detections or the inability to detect objects, including poor weather conditions, a decrease in the sensor field of view ~\cite{zhang_lidar_2021,dreissig_survey_2023}, and hardware-related issues such as blockages~\cite{vikas_camera_2022}. 
Predicting the movements of concealed road users is of increasing concern. Usually, there is no definite information about occluded objects which have not been observed before (unless communicated information is employed~\cite{buchholz_handling_2022,narri_set-membership_2021}). As a result, algorithms have been developed which produce reachable sets. Using these sets, possible planar positions of potentially hidden objects during the prediction stage are provided, similar to certain prediction methods for visible objects~\cite{althoff_set-based_2016,koschi_set-based_2020}. The works in~\cite{wang_reasoning_2021,nager_what_2019} propose an approach for producing reachable sets of hidden obstacles, incorporating historical information, while in~\cite{wang_reasoning_2021} it is assumed that these potentially hidden road users drive in the center of the lane. In~\cite{nager_what_2019}, however, the study is limited to pedestrians. The work in~\cite{sanchez_foresee_2022} does not adhere to these assumptions. In this case, concealed road users may emerge from any part of the obscured area, and it is assumed that they obey traffic regulations concerning the route and maximum velocity of the respective road. 

Incorporating potentially hidden road users in a planning strategy has led to different approaches. In~\cite{gilhuly_looking_2022,narksri_occlusion-aware_2022}, sampling-based planners are proposed which aim to gain more information over time, that is, reduce uncertainty on areas with potentially hidden road users, although this method disregards vehicle-dynamic information and constraints. The work in~\cite{firoozi_occlusion-aware_2022} considers planning using the reachable set of potentially hidden pedestrians as a collision constraint. Although recursive feasibility is guaranteed, progress along the path is not, as highly occluded scenarios could result in a complete blockage of the planner through constraints. Finally, the work in~\cite{zhang_safe_2021} considers trajectory planning under high occlusions as a dynamic game between the vehicle and traffic participants, although only single-agent interactions are considered. However, these planners lack the generality needed for multi-objective planning, i.e., planning a trajectory in the presence of objects, infrastructure, and possibly hidden objects, where each aspect can have a different weight (i.e., risk) inside the program.

Currently, there exists no trajectory planning method that considers static and dynamic objects, road infrastructure, potential hidden obstacles, and varying risk levels in a multi-objective context. This is a significant problem since imposing restrictions on all such entities may lead to an overly cautious approach that impedes progress toward the intended objective. Resolving this issue would enable a planning approach that can navigate complex urban environments, where proactive measures are necessary to prevent hazardous interactions with presently occluded road users. This work's main contributions, differing from current approaches, encompass:
\begin{enumerate}[label=(\roman*)]
    \item {\textit{Risk field construction:} We propose a framework for incorporating entities (e.g., objects, infrastructure) of arbitrary shape and size, into a continuous-time risk field. Starting from an occupancy grid, we exploit tools from the image processing literature and introduce the notion of \textit{risk kernels}. The convolution of an occupancy grid with the risk kernel constructs the risk field of the entity.}
    \item {\textit{Occlusion-awareness:} Using our proposed framework, we can incorporate present and future states of possible hidden obstacles, without assuming prior knowledge about their shape, orientation, and type (e.g, motorcycle or car). By including this level of situational comprehension, the planner can anticipate hidden obstacles and prevent safety-critical situations caused by, e.g., a limited field of view, degrading hardware, or fault-induced phenomena.}
    \item {\textit{Occlusion-aware trajectory generator:}
    Using the continuous-time risk field constructions, we propose a model-predictive trajectory generator based on minimizing a multi-objective cost comprising of risk and progress, allowing the trajectory to safely and efficiently maneuver in complex environments involving (potentially hidden) entities of arbitrary shapes.}
\end{enumerate} 
The outline of the remainder of this work is as follows. First, we introduce the problem statement and describe an outline of our proposed approach in Sec.~\ref{sec:problemformulation}. Subsequently, the theoretical framework is presented, which starts with the methodology to capture information on entities of different nature (objects, infrastructural elements and potentially hidden objects) into discrete occupancy grid maps in Sec.~\ref{sec:discretization}. These discretized descriptions are then converted into continuous risk fields using the \textit{risk kernel} description in Sec.~\ref{sec:riskfieldconstruction}. In Sec.~\ref{sec:mpcconstruction}, the proposed model predictive planning program is presented. Next, we test and prove the effectiveness of our proposed approach in a simulation study with several impactful scenarios involving, amongst others, vulnerable road users such as pedestrians and motorcycles in Sec.~\ref{sec:simulationresults}. Finally, Sec.~\ref{sec:conclusion} concludes the work.
\section{Problem Formulation and Proposed Approach}~\label{sec:problemformulation}

The problem of generating a feasible trajectory $\mathcal{T}$ from an initial state $\mathbf{x}_k$ towards a terminal state $\mathbf{x}_{k+N}$ using a model-predictive method involves enforcing a dynamic model as a constraint. The proposed approach in this work does not require the use of a particular dynamical model but does require the model to contain planar states $(x_k,y_k)$ to move across the Euclidean plane. As an example, a kinematic bicycle model is employed in Sec.~\ref{sec:simulationresults} to demonstrate the effectiveness of our approach. This model is highly suitable for low-speed urban environments as selected in the simulation study. It is ultimately up to the user of the algorithm to confidently choose the model that best suits their specific application domain.

The feasibility and efficacy of the trajectory $\mathcal{T}$ is further governed by the ability to progress towards a goal of interest, contained in a reference state $\mathbf{r}_{k+N}$, while minimizing the risk of, e.g., colliding with an object, unnecessarily crossing road markings, or neglecting potentially occluded objects. Similar to~\cite{van_der_ploeg_long_2022}, we assume that the reference state $\mathbf{r}_{k+N}$ is given by a global planner, which provides a low-resolution description of where the vehicle should drive towards and at which velocity. 

Besides the goal of reaching this reference state, risk mitigation is another (potentially conflicting) objective of the trajectory generator. Incorporating the presence of obstacles such as dynamic objects, occluded objects, and road infrastructural elements inside the trajectory generation problem, in order to support risk mitigation, poses significant challenges. 
Artificial potential fields (APFs) (also used and named as risk fields~\cite{van_der_ploeg_long_2022}) are used in literature to model and incorporate risk in the objective of a motion planner. Using the work in~\cite{jensen_mathematical_2022}, we define this risk as a measure to observe how close a vehicle's state is to violating a safety property. Safety properties can be, e.g., colliding with an obstacle, exiting the road, or hitting potentially hidden obstacles at a future time instance.  Using this approach, objects are, e.g., approximated by single points, located in the geometric center of the respective object, around which a deterministic bidirectional Gaussian repulsive field is constructed. This repulsive field results in collision-averse behavior when incorporated into the trajectory planning objective. Note that approaches exist in the literature that incorporate stochastic aspects related to the motion of other objects in the APFs (e.g., the stochastic reachability sets used in~\cite{malone_hybrid_2017}). In this study, we focus exclusively on a fully deterministic setting, prioritizing depth and clarity in our contributions. Further exploration in stochastic settings is considered future work. Road markings are approximated through polynomials~\cite{dixit_trajectory_2018,van_der_ploeg_long_2022}, around which a one-dimensional Gaussian repulsive field is constructed. This type of risk field results in enforcing the vehicle to stay within its lane and, potentially, stay in the center of the lane depending on the shape and size of the Gaussian risk fields. Finally, for polygonal shapes (e.g., vehicles or sidewalks), over-approximations of the shape with multiple bi-directional Gaussian fields are employed (e.g., through the use of Gaussian Mixture Models (GMM)~\cite{mok_gaussian-mixture_2017}), formed by a concatenation of convex shapes (i.e., ovals and circles) on a plane. The primary challenges associated with the above-mentioned approaches to constructing risk fields are:
\begin{enumerate}
\item \textit{Complex shapes and sizes of objects:} Approximating the object shape through a circular or ellipsoidal-shaped Gaussian potential field around a single point could lead to unsafe or conservative approximations of the (possibly non-convex) object shape by the risk field. For example, an under-approximation could result in trajectories that collide with the actual object, whereas an over-approximation could result in conservative trajectories (i.e., slow progression towards the goal), hindering the progress of the automated vehicle.
\item \textit{Non-smooth and discontinuous shapes of infrastructural elements:} Approximating edges of infrastructural elements, such as road markings or sidewalks, as polynomials could fail to capture the potentially non-smooth and discontinuous character of these elements in an urban setting. For example, road markings stop when traveling toward an intersection, which can't be captured using a continuous polynomial description. Moreover, in the same situation, sidewalks can take perpendicular turns which can't be captured with low-order polynomials. Approximating the aforementioned example phenomena through polynomials will give an inaccurate presentation of the real world and can hence result in trajectories that do not respect the actual road geometry, which, in turn, could induce unsafe situations.
\item \textit{Reachable sets of occluded objects:} Reachable sets of hidden objects, represented as polygons~\cite{sanchez_foresee_2022}, currently cannot be captured in a model-predictive planner without over-approximating the reachable sets of hidden objects through, e.g., GMMs. This can result in conservative behavior of the vehicle, hindering the progress of the AV.
\end{enumerate}
In this work, we consider three types of objects: static or dynamic objects in the field of view (e.g., vehicles and pedestrians), hidden objects (e.g., occluded vehicles and pedestrians) and infrastructural elements (e.g., road markings, sidewalks). We assume that we can measure the position and shape of static or dynamically moving objects within our field of view. Furthermore, we propose a framework to predict the positions of potentially hidden objects, building on the approach in~\cite{sanchez_foresee_2022}. Finally, we assume to be able to measure the position and shape of all relevant infrastructural elements of interest. Infrastructural elements prescribe the boundaries and the direction of the road, e.g. (but not limited to), road markings or sidewalks. All the gathered information can be mapped to the Euclidean plane $\set{E}\in\R^2$ and, if dynamic, is captured and predicted from current time up to $N\in\Z$ time steps ahead, where $N$ denotes the maximum prediction horizon. Consider that we can capture the space occupied by static and dynamic objects and future-time predictions hereof up to time instance $N$, i.e., $\set{O}_{k+n}^o,\:\forall o\in[1\hdots N_o],\:\forall n\in[0\hdots N]$ in a time-dependent collection $\set{O}_{k+n}$ of $N_o$ number of objects, such that $\set{O}_{k+n}=\{\set{O}_{k+n}^{1},\hdots,\set{O}_{k+n}^{N_o}\}\:\forall n\in[0\hdots N]$ 
and $\bigcup_{o=1}^{N_o} \set{O}_{k+n}\subseteq\set{E}\:\forall n\in[0\hdots N]$. Furthermore, consider that we can capture the space occupied by each infrastructural element by $\set{I}^i,\:\forall i\in[1\hdots N_i]$ and collect these observations of $N_i$ number of infrastructural elements in a time-invariant collection $\set{I}=\{\set{I}^{1},\hdots,\set{I}^{N_i}\}$, such that $\bigcup_{i=1}^{N_i}\set{I}\subseteq\set{E}$. Finally, consider that we can measure and predict the potentially occupied areas of hidden objects, i.e., $\set{H}_{k+n}^h,\:\forall h\in[1\hdots N_h],\:\forall n\in[0\hdots N]$ as time-varying collections $\set{H}_{k+n}=\{\set{H}^{1}_{k+n},\hdots,\set{H}^{N_h}_{k+n}\}\:\forall n\in[0\hdots N]$ of $N_h$ subsets, such that $\bigcup_{h=1}^{N_h}\set{H}_{k+n}\subseteq\set{E}\:\forall n\in[0\hdots N]$. The measurement and predictions of these areas occupied by hidden objects are based on a reachable set analysis of hidden objects and are further elaborated upon in Sec.~\ref{sec:hiddenobjects}. The variable $N_h$ denotes the number of classes the considered potentially hidden objects can exhibit, e.g., in the simulation study of this work in Sec.~\ref{sec:simulationresults} we differentiate two classes: vehicles and pedestrians (i.e., $N_h=2$). This allows us to, based on domain knowledge, treat certain areas differently than others based on the admissible inputs and states such objects may have in those respective areas. For example, a pedestrian can generally walk in any direction and change direction in a relatively short time. A vehicle generally follows the direction of the lane, and can't change its direction instantaneously.

\textbf{Problem statement: }The research question we aim to answer is how to generate a trajectory $\mathcal{T}$ towards a terminal reference goal $\mathbf{r}_{k+N}$, which incorporates all information from the collections $\set{O}_{k+n},\:\set{H}_{k+n}$ and $\set{I}$, to construct a trajectory which induces minimal risk while progressing towards $\mathbf{r}_{k+N}$. It is important to consider the shapes and sizes of all surrounding entities to construct risk fields while preserving the shape of the entities and preventing under or over-approximation, which could lead to risky or conservative solutions, respectively. Furthermore, we compare to a trajectory generator which does not incorporate any information about potential hidden traffic to understand the safety and comfort benefits of considering occluded objects. 
\begin{figure}[t]
    \centering
     \fontsize{9pt}{11pt}\selectfont% or whatever fontsize you like
    \def\svgwidth{1\columnwidth}
\begingroup%
  \makeatletter%
  \providecommand\color[2][]{%
    \errmessage{(Inkscape) Color is used for the text in Inkscape, but the package 'color.sty' is not loaded}%
    \renewcommand\color[2][]{}%
  }%
  \providecommand\transparent[1]{%
    \errmessage{(Inkscape) Transparency is used (non-zero) for the text in Inkscape, but the package 'transparent.sty' is not loaded}%
    \renewcommand\transparent[1]{}%
  }%
  \providecommand\rotatebox[2]{#2}%
  \newcommand*\fsize{\dimexpr\f@size pt\relax}%
  \newcommand*\lineheight[1]{\fontsize{\fsize}{#1\fsize}\selectfont}%
  \ifx\svgwidth\undefined%
    \setlength{\unitlength}{396.8503937bp}%
    \ifx\svgscale\undefined%
      \relax%
    \else%
      \setlength{\unitlength}{\unitlength * \real{\svgscale}}%
    \fi%
  \else%
    \setlength{\unitlength}{\svgwidth}%
  \fi%
  \global\let\svgwidth\undefined%
  \global\let\svgscale\undefined%
  \makeatother%
  \begin{picture}(1,0.71428571)%
    \lineheight{1}%
    \setlength\tabcolsep{0pt}%
    \put(0,0){\includegraphics[width=\unitlength,page=1]{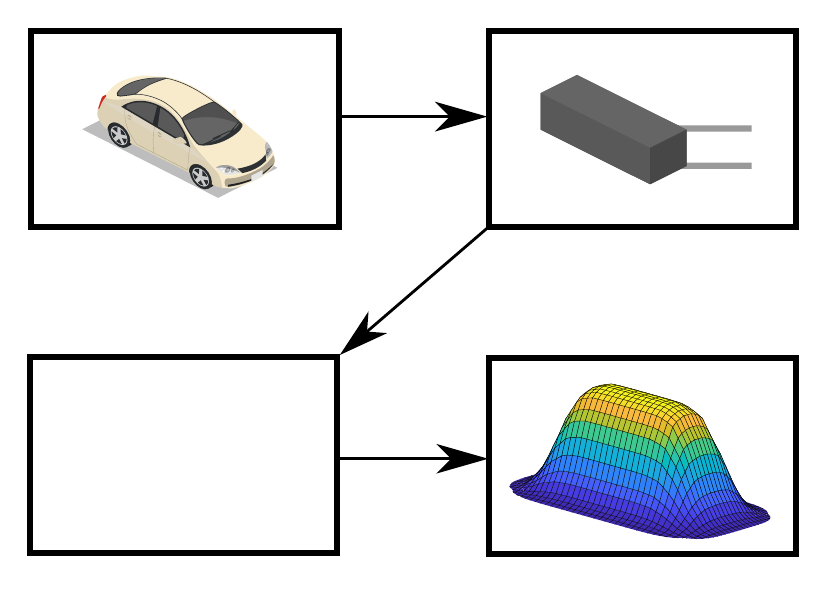}}%
    \put(0.06286417,0.40217495){\color[rgb]{0,0,0}\makebox(0,0)[lt]{\lineheight{1.25}\smash{\begin{tabular}[t]{l}Observation/prediction\end{tabular}}}}%
    \put(0.65257029,0.40548534){\color[rgb]{0,0,0}\makebox(0,0)[lt]{\lineheight{1.25}\smash{\begin{tabular}[t]{l}Occupancy grid\end{tabular}}}}%
    \put(0,0){\includegraphics[width=\unitlength,page=2]{fig2.pdf}}%
    \put(0.91459209,0.54581761){\color[rgb]{0,0,0}\makebox(0,0)[lt]{\lineheight{1.25}\smash{\begin{tabular}[t]{l}1\end{tabular}}}}%
    \put(0.91539819,0.49646105){\color[rgb]{0,0,0}\makebox(0,0)[lt]{\lineheight{1.25}\smash{\begin{tabular}[t]{l}0\end{tabular}}}}%
    \put(0.47560125,0.59509287){\color[rgb]{0,0,0}\makebox(0,0)[lt]{\lineheight{1.25}\smash{\begin{tabular}[t]{l}I\end{tabular}}}}%
    \put(0.52932178,0.33717703){\color[rgb]{0,0,0}\makebox(0,0)[lt]{\lineheight{1.25}\smash{\begin{tabular}[t]{l}II\end{tabular}}}}%
    \put(0.4756152,0.17748141){\color[rgb]{0,0,0}\makebox(0,0)[lt]{\lineheight{1.25}\smash{\begin{tabular}[t]{l}III\end{tabular}}}}%
    \put(0.07968527,0.00819996){\color[rgb]{0,0,0}\makebox(0,0)[lt]{\lineheight{1.25}\smash{\begin{tabular}[t]{l}Discrete risk field\end{tabular}}}}%
    \put(0.61272033,0.00797577){\color[rgb]{0,0,0}\makebox(0,0)[lt]{\lineheight{1.25}\smash{\begin{tabular}[t]{l}Continuous risk field\end{tabular}}}}%
  \end{picture}%
\endgroup%

    \caption{Three-step workflow from sensor observations towards a piecewise continuous risk field.}
    \label{fig:workflow}
\end{figure}

\textbf{Proposed approach:} our proposition entails a three-step workflow (Fig.~\ref{fig:workflow}) to convert discrete observations and predictions related to objects, hidden objects, and infrastructure into risk fields. In step I, we propose to map the shapes of interest (i.e., the collections $\set{O}_{k+n},\:\set{H}_{k+n}$ and $\set{I}$) into a discrete occupancy field. In step II, we convert this set of occupancy grids into a discrete potential field using a kernel-based technique. This technique allows a scalable and diverse approach for the construction of risk fields, as every entity can be assigned a \textit{risk kernel}, i.e., a distinct risk characteristic. In step III, we convert the discrete risk fields into a piecewise continuously differentiable risk field, constructed by a cubic spline surface. This conversion allows us to use artificial potential fields, also known as risk fields, to formulate objectives for model-predictive trajectory planning. The efficacy of our proposed method is illustrated by several impactful scenarios, as shown in Fig.~\ref{fig:problemstatement:icogram1}. In these scenarios, an automated vehicle (represented by blue color) encounters a complex urban environment where it could end up in a safety-critical situation with, e.g., a pedestrian appearing from between parked vehicles, a motorcyclist appearing from behind a vehicle at a crossing, or an approaching vehicle obstructed by vehicles/buildings. We provide this motivation to advocate the use of our method, which can anticipate and negotiate such situations through the scalable incorporation of dynamic objects, reachable sets of potentially hidden objects, and complex infrastructural shapes.

\section{Discretizing Perception and Predicting Hidden Objects}~\label{sec:discretization}
In this section, we aim to construct discrete occupancy maps which characterize the presence of entities so that we can follow the workflow shown in step I in Fig.~\ref{fig:workflow}. To map all the entity types relevant for the trajectory generation problem (i.e., objects, road markings, or potentially hidden objects), consider a discrete occupancy grid map with a cell resolution $r$, composed of cells with (integer) cell coordinates $(c_x,c_y)\in\Z$. 
A point $p = (p_x, p_y)\in\set{E}$ can be mapped to a cell coordinate using the following operation
\begin{equation}
    g(p_x, p_y) := (\underbrace{\left\lfloor p_x r + 0.5/r \right\rfloor}_{c_x}, \underbrace{\left\lfloor p_y r + 0.5/r \right\rfloor}_{c_y}).~\label{eq:cellresolution}
\end{equation}
A grid cell operator, $\opr{C}(\cdot)$, can collect a discrete set of cell coordinates contained within an arbitrary set $\exampleset\subseteq\set{E}$, using the set-builder notation, as follows:
\begin{align}
    \opr{C}(\exampleset) \!=\! \{ (c_x,c_y) \mid  \exists (x,y) \in \exampleset,\: g(x, y) = (c_x,c_y) \},
\end{align}
where the cell coordinates can take values in $\underline{c}_x\leq c_x\leq \bar{c}_x$ and $\underline{c}_y\leq c_y\leq \bar{c}_y$ with $\underline{c}_x,\bar{c}_x,\underline{c}_y,\bar{c}_y\in\Z$ such that $\underline{c}_x-\bar{c}_x=m-1,\: \underline{c}_y-\bar{c}_y=n-1$, where $m,\:n\in\mathbb{Z}^+$ denotes the number of possible values of $c_x,\:c_y$. Using the operator $\opr{C}(\cdot)$, an over-approximating grid map operator, $\opr{G}(\cdot)$, can be defined to construct an occupancy grid set for discrete grid cell coordinates, as follows:
\begin{figure}[t]
    \centering
     \fontsize{9pt}{11pt}\selectfont% or whatever fontsize you like
    \def\svgwidth{1\columnwidth}
\begingroup%
  \makeatletter%
  \providecommand\color[2][]{%
    \errmessage{(Inkscape) Color is used for the text in Inkscape, but the package 'color.sty' is not loaded}%
    \renewcommand\color[2][]{}%
  }%
  \providecommand\transparent[1]{%
    \errmessage{(Inkscape) Transparency is used (non-zero) for the text in Inkscape, but the package 'transparent.sty' is not loaded}%
    \renewcommand\transparent[1]{}%
  }%
  \providecommand\rotatebox[2]{#2}%
  \newcommand*\fsize{\dimexpr\f@size pt\relax}%
  \newcommand*\lineheight[1]{\fontsize{\fsize}{#1\fsize}\selectfont}%
  \ifx\svgwidth\undefined%
    \setlength{\unitlength}{269.29133858bp}%
    \ifx\svgscale\undefined%
      \relax%
    \else%
      \setlength{\unitlength}{\unitlength * \real{\svgscale}}%
    \fi%
  \else%
    \setlength{\unitlength}{\svgwidth}%
  \fi%
  \global\let\svgwidth\undefined%
  \global\let\svgscale\undefined%
  \makeatother%
  \begin{picture}(1,0.54736842)%
    \lineheight{1}%
    \setlength\tabcolsep{0pt}%
    \put(0,0){\includegraphics[width=\unitlength,page=1]{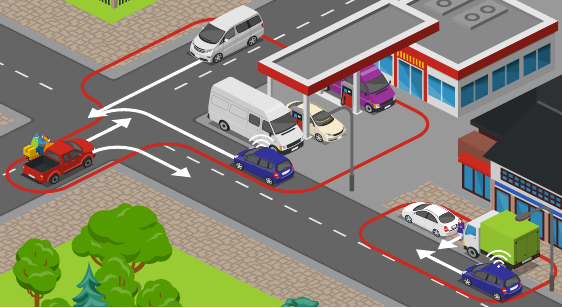}}%
  \end{picture}%
\endgroup%

    \caption{Several example cases for which an occluded view could cause a safety hazard.}
    \label{fig:problemstatement:icogram1}
\end{figure}
\begin{align}
    \opr{G}(\set{S})=&\{(p_x,p_y)\in\set{E}\mid g(p_x,p_y)\in\opr{C}(\exampleset)\}.
\end{align}

\subsection{Infrastructure measurements}
Assuming a description of all infrastructural elements in the vicinity of the vehicle to be available and measurable through, e.g., camera, LIDAR measurements, or high-definition maps, then each static element (e.g., road markings or sidewalks) can be considered a subset of the collection $\set{I}$. No restrictions apply to the shape and size of these elements, as long as these can be projected onto the Euclidean plane $\set{E}$ and could physically hinder the ego-vehicle. Now, considering the collection of infrastructural elements, $\set{I}$, we can capture their individual positions into discrete cell coordinates as:
\begin{align}
    \opr{C}(\set{I}^{i})=&\{ (c_x,c_y) \mid  \exists (x,y) \in \set{I}^{i},\nonumber\\&\: g(x, y) = (c_x,c_y) \},\:\forall i\in[1\hdots N_i],~\label{eq:occ_infrastructure}
\end{align}
such that all cells belonging to the position of an infrastructural element are occupied. In Sec.~\ref{sec:riskfieldconstruction}, this grid will be employed to capture the risk of crossing these infrastructural elements.
\subsection{Object measurements and predictions}
In this work, we assume that the shape and size of surrounding objects within the field of view (i.e., not obstructed), e.g., pedestrians or vehicles, are known or measurable at current time instance $k$. This assumption is fair since the literature offers effective methods for detecting, classifying, and estimating these shapes and sizes~\cite{ku_joint_2018}. Since motion prediction is not in the scope of this work, we assume to have access to the predictions of such objects from current time $k$, up to $N$ time steps ahead (i.e., toward time $k+N$). Note, however, that the literature is rich in various solutions for predicting the motion of vehicles and other road users (as collected and presented in~\cite{sanchez_scenario-based_2022}). In this work, we assume the measurements and predictions to be fully deterministic; however, in Sec.~\ref{sec:riskfieldconstruction} we will further elaborate on the incorporation of potential uncertainties surrounding these predictions. The measurements and predictions of objects are captured inside the collections $\set{O}_{k},\forall k\in[0\hdots N]$. Each set belonging to these collections can be characterized as a set of cell coordinates as follows:
\begin{align}
    \opr{C}&(\set{O}_{k+n}^{o})=\{ (c_x,c_y) \mid  \exists (x,y) \in \set{O}_{k+n}^{o}, g(x, y) = (c_x,c_y) \},\nonumber\\&\forall o\in[1\hdots N_o],\forall n\in[0\hdots N].~\label{eq:occ_object}
\end{align}
Similarly, for all $n\in[0\hdots N]$, occupancy grids can be constructed to capture the presence of other traffic participants over time. In Sec.~\ref{sec:riskfieldconstruction}, this grid will be employed to capture the risk that these observed objects pose for the host vehicle.

\subsection{Defining reachable sets of hidden objects}~\label{sec:hiddenobjects}
To construct a risk field from possible hidden objects, we build upon the work in \cite{sanchez_foresee_2022}. There, sets of states of possible hidden objects are computed using reachability analysis and sequential sensor measurements. In \cite{sanchez_foresee_2022}, free space measurements and sets of possible object states were represented as polygons. In this section, we describe how a discrete version of this algorithm can be implemented when the field of view and the computed sets of hidden objects are represented with occupancy grid maps.

First, we assume that objects, unseen at current time, of $N_h$ different classes (e.g., vehicles, pedestrians) are constrained to move within areas $\set{A}^{a,h} \subseteq \set{E},\:\forall h\in[1\hdots N_{h}],\:\forall a\in[1\hdots N_{a,h}]$, where $N_{a,h}$ represents the number of areas for each class $h$. For example, vehicles are assumed to only drive inside lanes and the movement of pedestrians is constrained to walkable areas (e.g., a pedestrian crossing or a sidewalk). In densely populated urban areas, pedestrians are assumed to have the freedom to move around anywhere. However, this assumption may not be realistic in highway situations, as this could lead to highly conservative behavior. To prevent this, designated areas where pedestrians are assumed to move around are identified in advance, such as through a map. It is important to note that these areas can overlap, such as a crossroad that can be used by both pedestrians and vehicles.

Next, we assume access to the currently detected free space $\set{F}_k\subseteq\set{E}$. This measurement can be obtained by multi- or single-sensor fusion, e.g., camera or LIDAR measurements and is limited by, for instance, sensor range or obstructions. Converting the set $\set{F}_k$ to an occupancy grid map using the operator $\opr{G}(\cdot)$ would result in an over-approximation of the free space. As a result, the area of potentially hidden objects would be under-approximated, which could result in safety-critical hazards.
Therefore, we base the further analysis on unobserved space, $\set{E}\setminus\set{F}_k$.

With the above assumptions, the occupancy grid map of possible hidden objects, at the initial time $k=0$, for any fixed given pair $(a,h)$, i.e., $\opr{G}(\set{H}^{a,h}_0)$ is
\begin{align}
    \opr{G}(\set{H}^{a,h}_0) &=\opr{G}(\set{E}\setminus\set{F}_0\cap \set{A}^{a,h}).\label{eq:G(H_0)^A}%\\&=\{ (c_x,c_y) \in& \Z \mid (c_x,c_y) \in \opr{G}(\set{F}^{\mathsf{c}}_0) \land (c_x,c_y) \in \set{A}^{a,h}\}.\label{eq:G(H_0)^A}
\end{align}
The interpretation is that, at the initial time, all cells that are not completely in our field of view (and hence are in the complement of the {\it detected} free space) and belong to an area where we assume hidden obstacles could exist, are in fact considered to be possibly occupied.

To reason about hidden objects at time steps $k > 0$, we need to assume object movement in the Euclidean plane at a discrete time step $k \in \mathbb{N}$ to follow a discrete-time dynamic map (modeling the motion of the hidden object):
\begin{equation}
    s_{k+1} = f(s_k, u_k)\label{eq:reachabilitydynamics}
\end{equation}
with positions $s_k, s_{k+1} \in \set{E}$ and input $u_k \in \set{U}$, where $\set{U}\subseteq \R^n$ denotes an admissible input set. 
If the admissible input set for an object depends on the tuple $({a,h})$, i.e., the area it is located in and the object class it belongs to, we denote it as $\set{U}^{a,h}$. For example, one could assume that vehicles only drive in the direction of the lane, i.e., they are not allowed to drive backward. 
The one-step reachable set for the dynamics in~\eqref{eq:reachabilitydynamics}, from an arbitrary initial set $\set{S} $ can now be defined as% a set of initial states $\set{X}^i$ can now be defined as
\begin{equation}
\label{eq:R(X^i)}
    \opr{R}(\exampleset) \coloneqq \{ f(s,u) \in \set{E} \mid s \in \exampleset \land u \in \set{U} \},
\end{equation}
and the one-step reachable set, constrained by an area $\set{A}^{(a,h)}$, is defined as
\begin{equation}
\label{eq:R^A(S)}
    \opr{R}(\exampleset\mid \set{A}^{a,h}) \coloneqq \{ f(s,u) \in \set{A}^{a,h} \mid s \in \exampleset \land u \in \set{U}^{a,h} \}.
\end{equation}

With the initial set $\opr{G}(\set{H}^{a,h}_0)$ from~\eqref{eq:G(H_0)^A} and the one-step reachable set, defined in~\eqref{eq:R^A(S)}, we can now over-approximate the set of possible hidden objects in an area $\set{A}^{a,h}$ at the time of planning, i.e., $n=0$, as
\begin{align}
    \set{H}_k^{a,h} = \{ (p_x, p_y) \in \opr{R}(\opr{G}&(\set{H}_{k-1}^{a,h})\mid \set{A}^{a,h}) \nonumber\\&\mid (p_x, p_y) \in \opr{G}(\set{E}\setminus\set{F}_k)\}.
\end{align}
Similarly to \cite{sanchez_foresee_2022}, we thus consider an area to be possibly occupied if 1) it is outside our current measured field of view $\set{F}_k$ and 2) was
 reachable by possible hidden objects outside our previous field view.
\begin{figure}[t]
    \centering
    \includegraphics[width=0.7\linewidth]{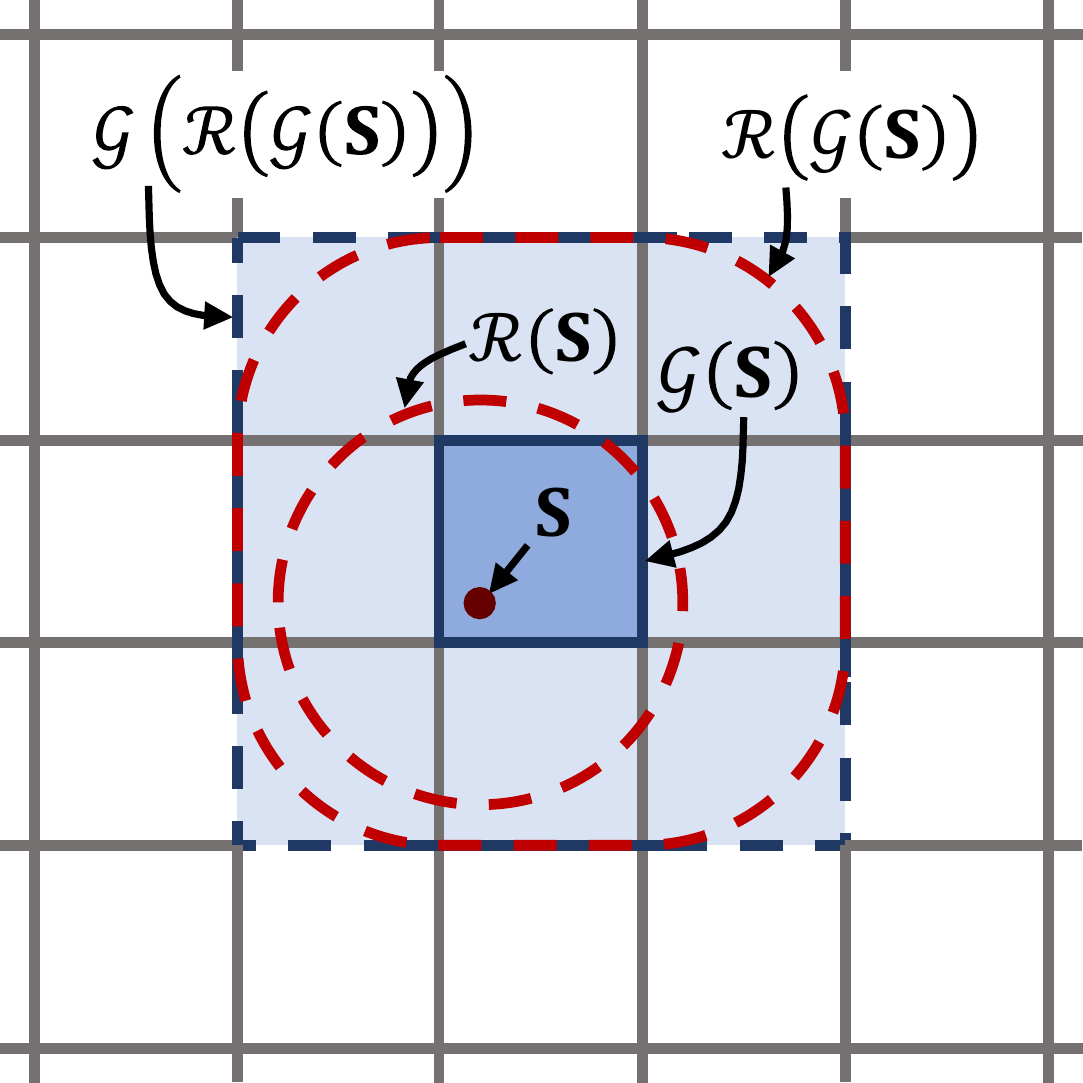}
    \caption{The reachable set $\opr{R}(\set{S})$ of a state $\set{S}$, the reachable set $\opr{R}(\opr{G}(\set{S}))$ of a grid cell $\opr{G}(\set{S})$, and the over-approximated grid representation $\opr{G}(\opr{R}(\opr{G}(\set{S})))$ of the reachable set of a grid cell.}
    \label{fig:grid_reachability}
\end{figure}

\subsection{Computing reachable sets of hidden objects}
Reachability analysis can be computationally intractable, even for relatively simple dynamics, $f(s,u)$. Therefore, in this work, we compute an over-approximation of the reachable set using a grid representation of the hidden obstacles.
Also, we consider a single discrete-time integrator model (i.e., $s_{k+1}=s_k+u_k$) for the dynamics in~\eqref{eq:reachabilitydynamics}, which with appropriate input constraints (i.e., using the maximum velocity a road user of a specific class can reasonably have) can over-approximate the underlying dynamics~\cite{koschi_set-based_2020}. Fig.~\ref{fig:grid_reachability} visualizes with dashed red lines the reachable sets $\opr{R}(\exampleset)$ and $\opr{R}(\opr{G}(\exampleset))$ of a single position $\exampleset$ and its over-approximated grid representation $\opr{G}(\exampleset)$ (here a single cell visualized in dark blue). The dashed blue line visualizes the border of the over-approximated grid representation $\opr{G}(\opr{R}(\opr{G}(\exampleset)))$, and the encompassed grid cells $\opr{G}(\opr{R}(\opr{G}(\exampleset)))$ are visualized in a lighter shade of blue.

The reachable set, $\opr{R}(\opr{G}(\set{S}))$, can be computed with a Minkowski sum with the admissible input set, $\set{U}$ \cite{hespanha_efficient_2006, althoff_computing_2022}, such that
\begin{equation}
    \opr{R}(\opr{G}(\set{S})) = \opr{G}(\set{S}) \oplus \set{U}.
\end{equation}
However, here we are interested in the reachable cells, $\opr{G}(\opr{R}(\opr{G}(\set{S})))$. Those can be computed with a discretized version of $\set{U}$, namely $\set{D}=\opr{G}(\set{U})$, such that
\begin{equation}
    \opr{G}(\opr{R}(\opr{G}(\set{S}))) = \opr{G}(\set{S}) \oplus \set{D}.\label{eq:structuringelement}
\end{equation}
\begin{Rem}
    The Minkowski sum, $\oplus$, with a discrete so-called \emph{structuring element} $D$ is typically referred to as a \emph{dilation}. This operation is fast to compute and readily available, for instance in MATLAB using the function \texttt{imdilate}.
\end{Rem}

To ensure that the computed reachable cells are within the area, $\set{A}^{a,h}$, the output of the dilation is intersected with $\set{A}^{a,h}$. This ensures that the reachable set of cells is over-approximated and efficiently computed as
\begin{equation}
    \label{eq:dilation}
    \opr{R}(\opr{G}(\set{H}_{k}^{a,h})\!\mid \!\set{A}^{a,h}) \subseteq \big( (\opr{G}(\set{H}_{k}^{a,h}) \oplus \set{D}^{a,h}) \cap \set{A}^{a,h} \big).
\end{equation}

For tracking possible hidden pedestrians, $\set{D}^{a,h}$ is an overapproximation of a disk-shaped element $\set{U}^{a,h}$ with a radius equal to their assumed maximum speed per time step. For vehicles, which are assumed to respect the driving direction of the lanes, the reachability is computed per road segment and $\set{U}^{a,h}$ is a semicircle-shaped structuring element oriented in the driving direction of the road segment, where $\set{D}^{a,h}$ is its overapproximation. The order of the computation of the road segments must follow the driving direction of the lane to allow vehicles to traverse to the next road segment. Following this assumption, we over-approximate the set of possible hidden obstacles in an area $\set{A}^{a,h}$ at time $k$ (i.e., $n=0$) as
\begin{align}
    \set{H}_k^{a,h} = \{ (p_x, p_y) \in & \big( (\opr{G}(\set{H}_{k-1}^{a,h}) \oplus \set{D}^{a,h}) \cap \set{A}^{a,h} \big) \nonumber\\&\mid (p_x, p_y) \in \opr{G}(\set{E}\setminus\set{F}_k)\}.
\end{align}

\begin{figure}[t]
    \centering
    \includegraphics[width=1\linewidth]{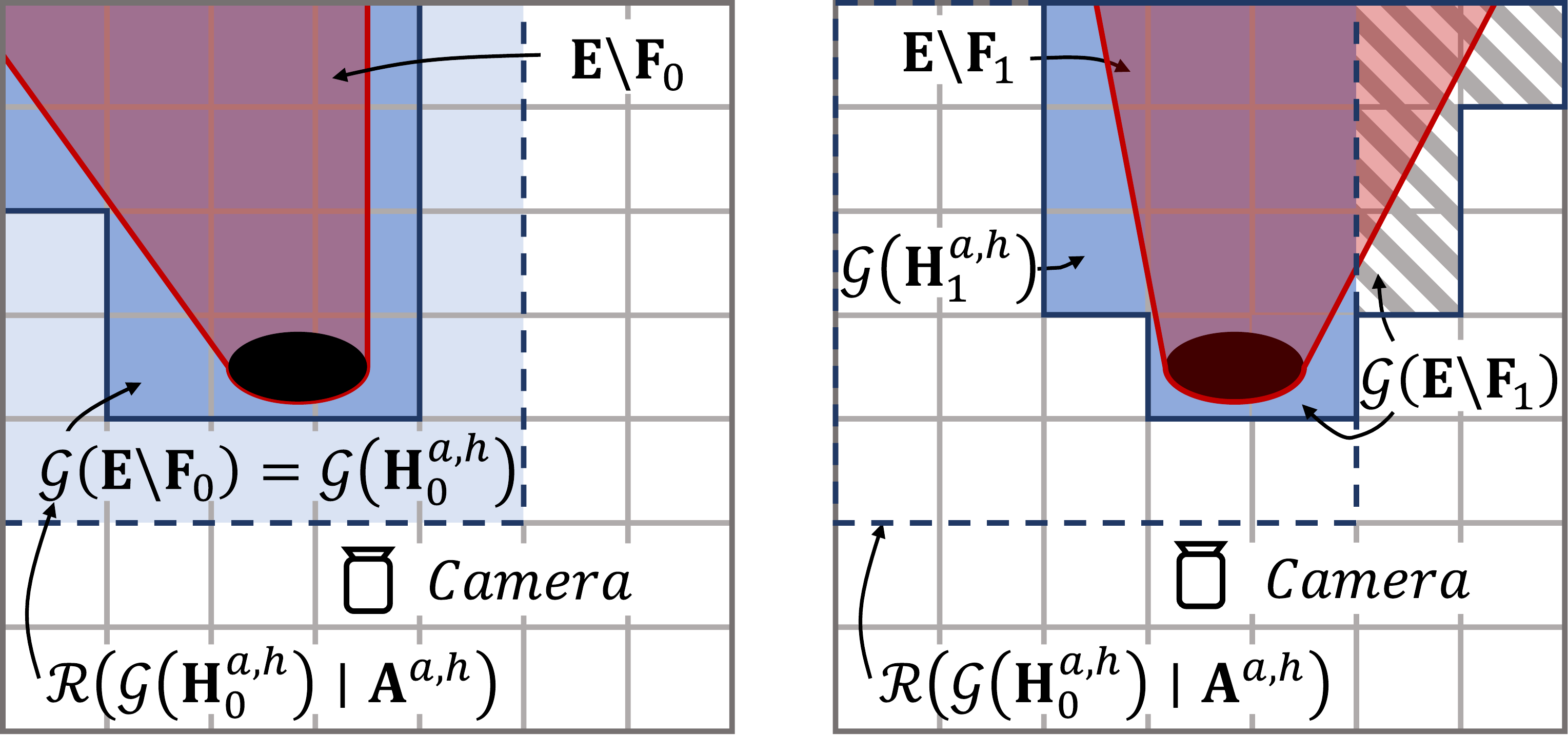}
    \caption{The reachable set of a state, the reachable set of a grid cell, and the over-approximated grid representation of the reachable set of a grid cell. Note that the red area is outside the camera's field of view.}
    \label{fig:method}
\end{figure}

An example of a result of this algorithm is shown in Fig.~\ref{fig:method}. For visualization purposes, we assume that there is only one area, $\set{A}$, that extends for the size of the map.
Fig.~\ref{fig:method}.a depicts an obstacle in black, causing a reduced detected free space $\set{F}_0$ observed from the camera location. At this first time step, the occupancy grid map of all the possible hidden obstacles $\opr{G}(\set{H}_0)$ is initialized as the over-approximated grid map of the complement of the detected free space $\opr{G}(\set{E}\setminus\set{F}_0)$. The one-step reachable set is shown as $\opr{R}(\opr{G}(\set{H}_0)\mid \set{A})$.
In Fig.~\ref{fig:method}.b, the obstacle has moved to the right, also causing a change in the detected free space $\set{F}_1$. At this time, the possible hidden obstacles $\opr{G}(\set{H}_1)$ must be both, outside the currently detected free space $\opr{G}(\set{F}_1)$, and within the reachable set of the possible hidden obstacles at the previous time step $\opr{R}(\opr{G}(\set{H}_0)\mid \set{A})$.

The set of possible hidden obstacles at a future time step, $k+n$, can be computed in a similar way, with the difference that no free space detection is available yet. Iteratively, the set of hidden obstacles can thus be computed as
\begin{align}
    \set{H}_{k+n}^{a,h} = \{ (p_x, p_y) \in \big( (\opr{G}(\set{H}_{k+n-1}^{a,h}) \oplus \set{D}^{a,h}) \cap \set{A}^{a,h} \big) \}.
\end{align}
with $n=[1, \ldots, N]$.
As mentioned before, the tracking of possible hidden obstacles is done per area, without hidden objects being able to cross from one area to another. For example, when a hidden (vehicle) object approaches a four-way intersection, it can take a left turn, a right turn, or proceed straight. Each of these areas to travel along is treated separately and may partially overlap. As such, the final step needed before giving these sets to the planner is to compute the union over the areas as
\begin{align}
    \set{H}_{k+n}^h=\bigcup_{a=1}^{
N_a}\set{H}_{k+n}^{a,h},\:\forall n\in[0\hdots N],\:\forall h\in[1\hdots N_h].~\label{eq:occ_hidden}
\end{align}
\section{Constructing the Risk Fields}\label{sec:riskfieldconstruction}
In this section, we introduce steps II and III (Fig.~\ref{fig:workflow}) of our method of converting the set representations of occupancies of different entities (treated in the previous section) into a continuous risk field to be incorporated into a model-predictive planning program. As a first step, a discrete occupancy grid, $\opr{C}(\exampleset)$, can be converted into an occupancy matrix, using the matrix operator $\opr{M}(\cdot)$, which element-wise constructs the occupancy matrix as follows: 
\begin{equation*}
\opr{M}(\exampleset) =
\begin{cases}
1, & \text{if } (c_x,c_y) \in \opr{C}(\exampleset). \\ 0, & \text{otherwise}.
\end{cases}
\end{equation*}
The matrix operation results in an occupancy grid matrix $\opr{M}(\exampleset)\in\R^{m\times w}$. By representing the occupancy of objects, infrastructural elements, or locations of possible hidden objects, we pave the way for incorporating the risk kernel.
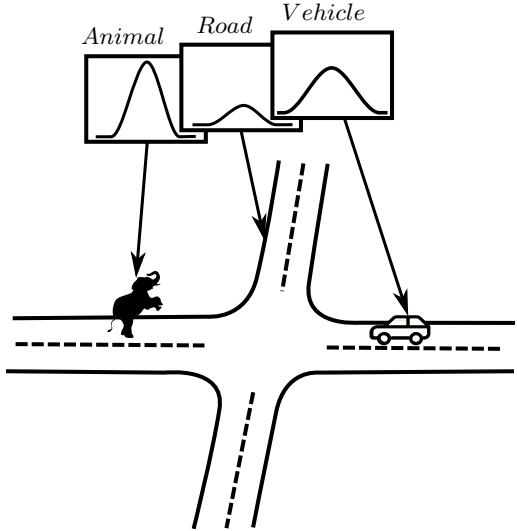
\begin{figure}[t]
    \centering
     \fontsize{9pt}{11pt}\selectfont% or whatever fontsize you like
    \def\svgwidth{0.8\columnwidth}
\begingroup%
  \makeatletter%
  \providecommand\color[2][]{%
    \errmessage{(Inkscape) Color is used for the text in Inkscape, but the package 'color.sty' is not loaded}%
    \renewcommand\color[2][]{}%
  }%
  \providecommand\transparent[1]{%
    \errmessage{(Inkscape) Transparency is used (non-zero) for the text in Inkscape, but the package 'transparent.sty' is not loaded}%
    \renewcommand\transparent[1]{}%
  }%
  \providecommand\rotatebox[2]{#2}%
  \newcommand*\fsize{\dimexpr\f@size pt\relax}%
  \newcommand*\lineheight[1]{\fontsize{\fsize}{#1\fsize}\selectfont}%
  \ifx\svgwidth\undefined%
    \setlength{\unitlength}{396.8503937bp}%
    \ifx\svgscale\undefined%
      \relax%
    \else%
      \setlength{\unitlength}{\unitlength * \real{\svgscale}}%
    \fi%
  \else%
    \setlength{\unitlength}{\svgwidth}%
  \fi%
  \global\let\svgwidth\undefined%
  \global\let\svgscale\undefined%
  \makeatother%
  \begin{picture}(1,1.05)%
    \lineheight{1}%
    \setlength\tabcolsep{0pt}%
    \put(0,0){\includegraphics[width=\unitlength,page=1]{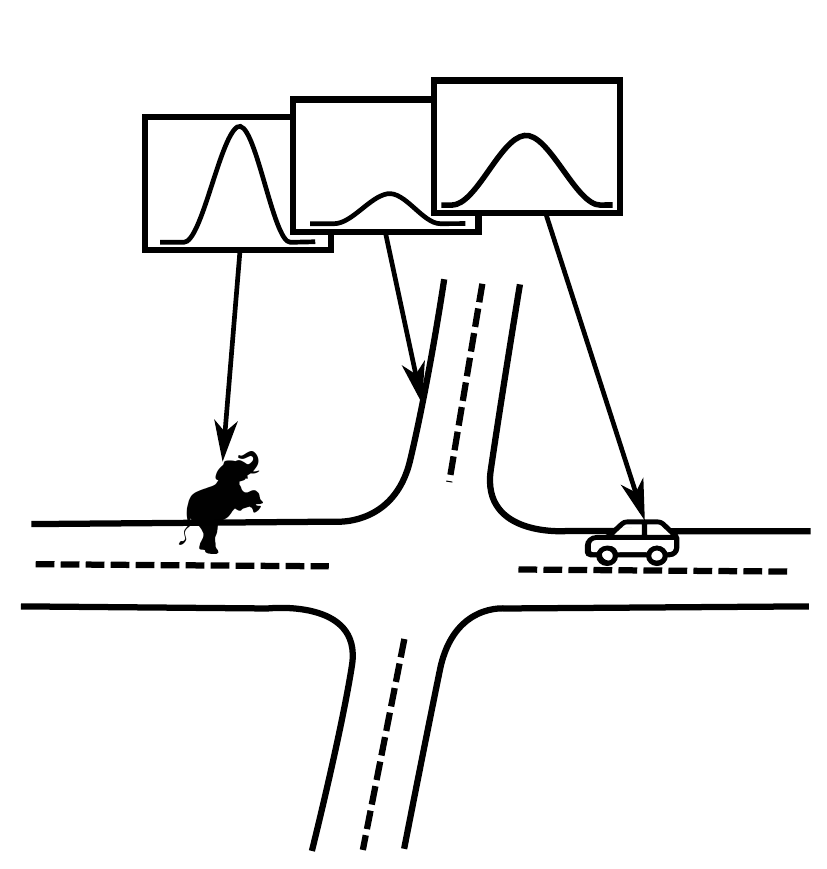}}%
    \put(0.16630933,0.93093774){\color[rgb]{0,0,0}\makebox(0,0)[lt]{\lineheight{1.25}\smash{\begin{tabular}[t]{l}$Animal$\end{tabular}}}}%
    \put(0.38243899,0.95198116){\color[rgb]{0,0,0}\makebox(0,0)[lt]{\lineheight{1.25}\smash{\begin{tabular}[t]{l}$Road$\end{tabular}}}}%
    \put(0.54307883,0.97654956){\color[rgb]{0,0,0}\makebox(0,0)[lt]{\lineheight{1.25}\smash{\begin{tabular}[t]{l}$Vehicle$\end{tabular}}}}%
  \end{picture}%
\endgroup%

    \caption{The concept of the risk kernel.}
    \label{fig:constructingriskfields:riskfieldconcept}
\end{figure}
\subsection{The risk kernel}
The main concept of the risk field is depicted in Fig.~\ref{fig:constructingriskfields:riskfieldconcept}. That is, each and every entity around the automated vehicle can induce a certain level of risk to be considered in the model-predictive planner. The risk kernel filters the occupancy-grid matrix, analogous to how kernels or masks are being used in image processing, e.g., blurring or sharpening of images. Through this approach, miscellaneous properties of objects, lines, and possible locations of hidden objects, related to the risk of harm, can be denoted through repulsive fields. 
In this work, we select the base function of the risk kernel to be a typical bivariate Gaussian function which is the most commonly used base function in literature to model risk fields. The parameters of this base function can differ for all $N_o$ objects, $N_i$ infrastructural elements, and $N_h$ different classes of possible hidden objects. As such, the risk kernel could be a function of the intrinsic properties of the considered entity, as demonstrated in earlier work~\cite{van_der_ploeg_connecting_2023}. For example, a solid road marking could exhibit a substantially higher risk due to its illegality to cross, when compared to a dashed marking, which is legal to cross. When considering objects, colliding with a small object (e.g., a cardboard box) objectively has a much lower risk for the automated vehicle than colliding with a car or a vulnerable road user. Therefore, one can state that the risk of harm is substantially higher for the latter case. When considering areas of potentially hidden objects, one could assign a different risk of harm to each class of potentially hidden objects. As an example, the risk of harm when colliding with a vulnerable road user within an urban environment is usually greater than the risk of harm when colliding with another vehicle. Moreover, vulnerable road users are more difficult to predict~\cite{sanchez_scenario-based_2022}. As such, the uncertainty related to the motion of these road users is higher, which, in the deterministic setting we consider, can be reflected by making the risk-kernel wider, by linking its standard deviation to the state-uncertainty provided by, e.g., Kalman filters~\cite{mora_path_2008}.

In order to characterize the risk of an entity using the occupancy of objects, denoted in Sec.~\ref{sec:discretization}, we introduce the notion of the risk kernel. Such a risk kernel provides the ability to describe the risk of harm related to an object, and the positional decay of this risk. Subsequently, using a discrete convolution of the risk kernel with our developed occupancy maps, the risk field of an entity can be constructed.

A risk mapping $k(z_1,z_2):(z_1,z_2)\in\mathbb{R}^{2\times1}\mapsto \mathbb{R}$ is exploited for the element-wise sampling of an admissible risk kernel matrix. A candidate mapping, considered in this work, is given as follows:
\begin{align}
    k(z_1,z_2)=ae^{-\frac{z_1^2+z_2^2}{2\sigma^2}}.~\label{eq:risk_function}
\end{align}
Note, that this mapping provides the sampling equivalent of the risk field employed in~\cite{van_der_ploeg_long_2022}, where $\sigma$ denotes the standard deviation of the Gaussian field and $a$ represents its magnitude.
By applying the mapping $k$ on the elements from coordinate vectors $\tilde{x}\in\mathbb{R}^{1\times b}$ and $\:\tilde{y}\in\mathbb{R}^{1\times p}$ (denoting evenly spaced vectors, centered around $(0,0)$), a kernel matrix $K\in\mathbb{R}^{b\times p}$ can be constructed as follows: 
\begin{align}
    K=\begin{bmatrix}k(\tilde{x}_{1},\tilde{y}_{1})&\hdots &k(\tilde{x}_{1},\tilde{y}_{p})\\\vdots&\ddots&\vdots\\k(\tilde{x}_{b},\tilde{y}_{1})&\hdots&k(\tilde{x}_{b},\tilde{y}_{p})\end{bmatrix}.~\label{eq:risk_kernel}
\end{align}
 The dimensions, $b,\:p$, and the distance between the coordinates in~$\tilde{x},\tilde{y}$ should be chosen in such a way as to sufficiently capture the shape of the Gaussian field. For example, one could cover $99.7\%$ of the Gaussian distribution by sampling from $-3\sigma$ to $3\sigma$ in both dimensions.  The resulting discrete kernel matrix $K$ can be employed to proceed with our proposed approach. 
\begin{Def}[Discrete risk field]
A discrete risk field (DRF) matrix can be constructed through the following convolution operation:
\begin{align}
    \opr{D}(\exampleset)=&K*\opr{M}(\exampleset).
\end{align}
To further elaborate on this operation, let us introduce the shorthand notations $\opr{D}_\exampleset(\cdot,\cdot),K(\cdot,\cdot),\opr{M}_\exampleset(\cdot,\cdot)$ for accessing the entries of the respective matrices $\opr{D}(\exampleset),K,\opr{M}(\exampleset)$. Then, for each entry $\opr{D}_\exampleset(d_x,d_y)$, the following operation is performed
\begin{align}
    \opr{D}_\exampleset(d_x,d_y)=&\nonumber\\\sum_{k_x}\sum_{k_y}\!K&(k_x,k_y)\opr{M}_\exampleset(d_x-k_x+\!1,d_y-k_y+\!1),
\end{align}
where $k_x,k_y$ range over all legal subscripts of $K$ and $\opr{M}_\exampleset$, respectively.
The resulting matrix $\opr{D}(\set{S})\in\R^{(m+b-1)\times (w+p-1)}$ forms a DRF representation of the set $\exampleset$.    
\end{Def}
\subsection{From discrete to continuous risk fields}
In order to construct a continuous description of the discrete risk field $F(\exampleset)$, which we can employ in a model-predictive control framework (which typically uses a gradient-based solver), a bicubic surface is generated. Using a cubic function as a basis allows us to exploit several advantageous properties. First, these functions are continuously differentiable between the data points in $\opr{D}(\exampleset)$ for which these cubic functions are constructed. Furthermore, these functions span the data points contained in the grid $\opr{D}(\exampleset)$ of interest, therefore accurately representing the desired discrete risk description~\cite{hou_cubic_1978}.  
Since the construction and determination of cubic splines are not new and can be found commonly in the literature (often used in, e.g., image processing), we make use of an existing method from literature and refer the reader to~\cite{de_boor_bicubic_1962} for the exact details on determining such a bicubic surface from a data set. The "\textit{not-a-knot}"~\cite{de_boor_convergence_1985} end conditions are used to make the linear problem of finding a bicubic surface consistent but also provide us a unique solution. 
\begin{Def}[Continuous risk field]
Let the tuple $(X,Y)$ represent a set of coordinates belonging to the Euclidean plane $\mathbf{E}$.
A continuous risk field (CRF), denoted by, $\opr{F}(\exampleset, X, Y)$, as a function of a DRF $\opr{D}(\exampleset)$, represents a mapping $(X,Y)\mapsto r$, where $r$ represents the risk magnitude. This mapping exists for all $(X,Y)\in\exampleset$. The CRF is continuous in all its first (mixed) partial derivatives.
\end{Def}

\section{Construction of the model-predictive planning program}~\label{sec:mpcconstruction}
By capturing all entities of interest as piecewise continuous risk field descriptions, the optimization program fulfilling our original problem statement can be constructed. As a basis, we take the program described in~\cite[Eq. 7]{van_der_ploeg_long_2022} as a basis and exclude the risk fields constructed around points for objects and polynomials for infrastructural elements, as employed in~\cite{van_der_ploeg_long_2022}. Instead, we augment the problem with our novel continuous risk-field descriptions as follows:
\begin{subequations}
\label{eq:mpcprogram}
    \begin{align}
        &J^\star_{0\rightarrow N}(\mathbf{x}_s)=\min_{u,x}\sum_{n=1}^{N-1}\ell_\mathbf{x}\left(\mathbf{x}_{k+n}-\mathbf{r}_{k+n}\right)+\sum_{n=0}^{N-1}\ell_\mathbf{u}\left(\mathbf{u}_{k+n}\right)\nonumber\\&+m(\mathbf{x}_{k+N}-\mathbf{r}_{k+N})+\sum_{n=1}^{N}\Big(\sum_{i=1}^{N_i}\opr{F}(\set{I}^{i},x_{k+n},y_{k+n})\nonumber\\&\sum_{o=1}^{N_o}\opr{F}(\set{O}_{k+n}^{o},x_{k+n},y_{k+n})+\sum_{h=1}^{N_h}\opr{F}(\set{H}_{k+n}^{h},x_{k+n},y_{k+n})\Big)\label{eq:pathmpc1}\\&\text{s.t.}\quad\:\:\mathbf{x}_k=\mathbf{x}_s\label{eq:pathmpc2}\\
        &\:\:\:\quad \mathbf{x}_{k+n+1}=f(\mathbf{x}_{k+n},\mathbf{u}_{k+n}),\quad\forall n\in[0\hdots N],\label{eq:pathmpc3}\\
        \quad& g(\mathbf{x}_{k+n},\mathbf{u}_{k+n})\leq\: 0,\quad\forall n\in[0\hdots N], \label{eq:pathmpc4}
    \end{align}
\end{subequations}
where, in~\eqref{eq:pathmpc1}, the functions $\ell_{\mathbf{x}}(\cdot),\:\ell_{\mathbf{u}}(\cdot)$ quadratically penalize the state error $\mathbf{x}_{k+n}-\mathbf{r}_{k+n}$ with respect to a reference state $\mathbf{r}_{k+n}$, and the input $\mathbf{u}_{k+n}$, respectively. Furthermore, the term $m(\cdot)$ quadratically penalizes the terminal state at time $N$. Finally,~\eqref{eq:pathmpc1} contains the continuous risk field descriptions of infrastructural elements~\eqref{eq:occ_infrastructure}, static/dynamic objects~\eqref{eq:occ_object}, and potentially hidden objects~\eqref{eq:occ_hidden}. The state $\mathbf{x}_s$ in~\eqref{eq:pathmpc2} represents the state at present time, i.e., $k=0$. The function $f$ in~\eqref{eq:pathmpc3} represents the non-linear dynamics as depicted in~\eqref{eq:vehiclemodel}. Finally, the constraints in~\eqref{eq:pathmpc4} enforce the bounds on the states in $\mathbf{x}$ and the inputs in $\mathbf{u}$.

\section{Simulation results}~\label{sec:simulationresults}\
In this section, we aim to demonstrate the contributions of this work in several scenarios, inspired by the original motivating examples from Fig.~\ref{fig:problemstatement:icogram1}. In short, the scenarios are summarized, as follows:
\begin{enumerate}
    \item \textit{Hidden vehicle:} In this scenario the ego-vehicle approaches an intersection, aiming to take a left turn. A building occludes the right-hand side of the intersection. Several vehicles travel along this intersection, one of which is inside the occluded area and is also traveling towards the intersection.
    \item \textit{Hidden pedestrian:} In this scenario, the ego-vehicle travels on a straight road along a strip of parked cars, e.g., next to a supermarket. A pedestrian, coming from the supermarket, is walking towards and through the occluded space between the parked cars to cross the road.
    \item \textit{Hidden motorcycle:} In this scenario, the ego vehicle approaches an intersection, to take a left turn. A truck is coming from the left which will take a right turn. Next to the truck, a motorcycle is traveling which intends to drive straight across the intersection. 
\end{enumerate}
The impact of these scenarios is illustrated by recent studies. For example, cases of crossing pedestrians with sight obstruction form a large percentage (up to around $33\%$) of fatal accidents with pedestrians in Germany over the past two decades~\cite{noauthor_safe-up_2021}. Furthermore, a recently released study shows that, for the motorcycle crossing scenario, around $30\%$ of the accidents in these cases have occurred in the presence of visual obstructions~\cite{noauthor_cmc_2021}. Using these vulnerable road user cases as illustrating cases, we depict the strength of our contributions as the planner anticipates these possibly hidden road users and preemptively takes action before possible dangerous situations occur. In the next section, we will elaborate on the simulation setup, after which the simulation results will be provided. All the figures and full videos of the considered scenarios are available at ${\tt github.com/tno-ivs/FearoftheDark}$.
\subsection{Simulation Setup}
As elaborated upon in Sec.~\ref{sec:mpcconstruction}, the model-predictive planner uses an internal model to predict the vehicle states for an optimal minimum-risk trajectory. The model used within the scope of this simulation study is a kinematic bicycle model, for which the dynamical discrete difference equations read as follows:
\begin{align}
\begin{aligned}
    x_{k+1} =& x_k+t_sv_k\cos{\theta_k},\quad y_{k+1} = y_k+t_sv_k\sin{\theta_k},\\
    \theta_{k+1} =& \theta_k+t_s\frac{v_k}{L}\tan{\delta_k},\quad v_{k+1} = v_k+t_sa_k,\\
    \delta_{k+1} =& \delta_k+t_s\omega_k,
\end{aligned}\label{eq:vehiclemodel}
\end{align}
such that the state vector is given as $\mathbf{x}_k=\{x_k,y_k,\theta_k,v_k,\delta_k\}$, and the input vector is given as $\mathbf{u}_k=\{a_k,\omega_k\}$. The symbol $k$ denotes the discrete-time counter, the states $x,y$ denote the planar position of the vehicle, the state $\theta$ denotes the heading, $v$ denotes the longitudinal velocity, $\delta$ denotes the steering angle, $a$ denotes the input longitudinal acceleration, $\omega$ denotes the input steering-rate and $L$ denotes the wheelbase of the vehicle. A subset pf the states are constrained by $[\underline{v},\underline{\delta}]\leq[v_k,\delta_k]\leq[\widebar{v},\widebar{\delta}]$ and the inputs are constrained by $[\underline{a},\underline{\omega}]\leq\mathbf{u}_k\leq[\widebar{a},\widebar{\omega}]$ for all time instances $k\in[k\hdots k+N]$. In the scope of this simulation study, we consider that the objects, infrastructural elements, and possibly hidden objects each have one risk field parametrization of the risk kernel in~\eqref{eq:risk_function}, i.e., $[a_o,\sigma_o],[a_i,\sigma_i],[a_h,\sigma_h]$ representing the amplitude and standard deviation of the risk kernel for each class respectively. Note that, for the sake of simplicity, but without loss of generality, we use a square discrete kernel with evenly spaced vectors $\tilde{x},\:\tilde{y}$ (as used in~\eqref{eq:risk_kernel}), such that $b=p$. 
Moreover, each class has its own distinct cell resolution (as used in~\eqref{eq:cellresolution}), i.e., $r_o, r_i, r_h$. We assume that the ego vehicle has a LiDAR sensor with a sensing radius of $100$m, sensing all objects around the ego-vehicle and within the field of view (that is, within LiDAR range and not being occluded). Each scenario runs with a hidden object class of vehicles (i.e., cars or motorcycles), where we assume that the vehicles travel in the direction of the lane, i.e., the structuring element $\set{D}^{a,h}$~\eqref{eq:structuringelement} is an overapproximation of $\set{U}^{a,h}$, which is a semi-circle-shaped element with a radius of $v_o\cdot t_s$. All the parameters mentioned in this section are provided in  Table~\ref{table:pars} and are considered constant throughout all simulations. All remaining parameters which are specific to the simulations themselves, e.g., the considered object classes and the initial conditions of the simulation are provided in their respective sections.

For the simulation studies, the model for simulation is identical to the internal model of the MPC, i.e., the dynamical equations given in~\eqref{eq:vehiclemodel}. The road layout used within this simulation study is from the Carla~\cite{dosovitskiy_carla_2017} \textit{Town03} OpenDrive file in an attempt to represent a naturalistic intersection layout. The model-predictive planner and the simulation are run in MATLAB, using CasADi~\cite{andersson_casadi_2019} for formulating the optimization program and IPOPT~\cite{wachter_implementation_2006} for solving the program using an interior-point method.

\begin{table}
\caption{Simulation parameters}
\label{table:pars}
\small
\begin{tabular}{p{1cm} p{3.35cm} p{2.1cm} p{0.6cm}}\toprule
\textbf{Par.}              & \textbf{Description} & \textbf{Value} & \textbf{Unit}\\
\toprule
$t_s$ & Sampling time & 0.4 & s\\
$N$            & Trajectory horizon & 10 & -\\
$\ell_\mathbf{u}$           & Input weight & $[0.5,1000]$ & -\\
$\ell_\mathbf{x}$ & Stage weight & $[0,0,0,0.1,0.1]$& -\\
$m$&Terminal weight&$[20,20,1,0.1,0.1]$&-\\
$\underline{a},\overline{a}$&Long. acceleration bounds&$[-5,2]$&$\text{ms}^{-2}$\\
$\underline{\delta},\overline{\delta}$&Steering angle bounds&$[-1.5,1.5]$&$\text{rad}$\\
$\underline{v},\overline{v}$&Long. velocity bounds&$[0,5]$&$\text{ms}^{-1}$\\
$\underline{\omega},\overline{w}$&Steering rate bounds &$[-1.5,1.5]$&$\text{rads}^{-1}$\\
$a_o,a_i,a_h$&Risk field amplitude&1300,\:200,\:110&-\\
$\sigma_o,\sigma_i,\sigma_h$&Risk field std. deviation&0.36,\:0.25,\:0.2&m\\
$r_o,r_i,r_h$&Grid resolution&0.2,\:0.2,\:0.2&-\\
\toprule
\end{tabular}
\end{table}
\begin{figure}
    \centering
    \includegraphics[width=1\linewidth]{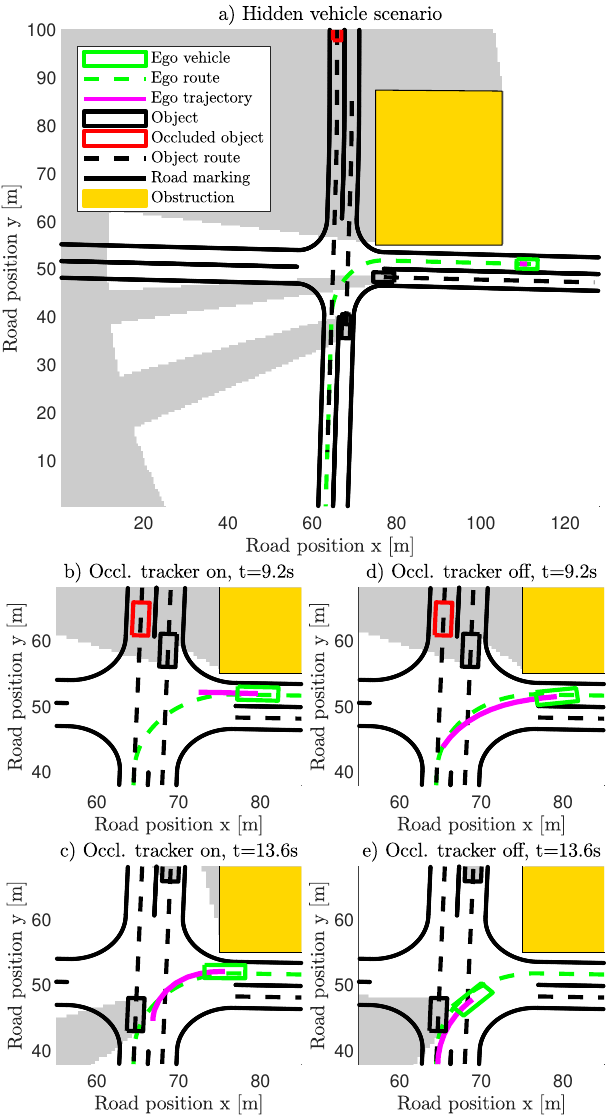}
    \caption{Overview of the "hidden vehicle" scenario and snapshots at several time instances for motion planning results with and without occlusion tracking.}
    \label{fig:sim:scenario_0_overview}
\end{figure}
\begin{figure}
    \centering
    \includegraphics[width=1\linewidth]{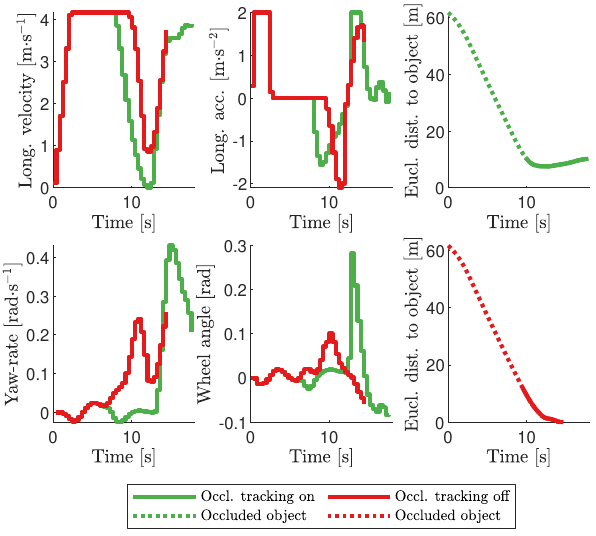}
    \caption{Signal plots for the "hidden vehicle" scenario.}
    \label{fig:sim:scenario_0_plots}
\end{figure}
\subsection{Scenario 1: Hidden vehicle}
The overall setup of the first scenario, where a hidden vehicle approaches the same intersection as the ego-vehicle, is given in Fig.~\ref{fig:sim:scenario_0_overview}a. We consider a single object class (i.e., $N_h=1$), where the considered maximum velocity of any hidden object is assumed to be~$v_o=4.44\text{m}\cdot\text{s}^{-1}$. The considered class is that of vehicles. Three vehicles populate this simulation, of which one is hidden (i.e., the one (in red) coming from the top of the intersection). The three non-ego vehicles, starting from the top and counting counter-clockwise have a velocity of $4.2,\:4.2$ and $2.2\text{m}\cdot\text{s}^{-1}$, respectively. Until $8$s,  (with and without occlusion tracking) behave identically as they accelerate from a standstill and approach the intersection. A small steering movement can be observed from Fig.~\ref{fig:sim:scenario_0_plots}, as both planners slightly evade a vehicle approaching from the opposite lane. We proceed our analysis with Fig.~\ref{fig:sim:scenario_0_overview}b and Fig.~\ref{fig:sim:scenario_0_overview}c, in combination with Fig.~\ref{fig:sim:scenario_0_plots}, where we observe the scenario at $9.2$s after the start of the simulation. In one sampling instance after this time, the occluded vehicle will be detected by the ego vehicle equipped with the planner without occlusion tracking. In one sampling instance after that, the vehicle will be detected by the planner with occlusion tracking, since the vehicle was already anticipating by braking and for that reason will observe the occluded vehicle at a later time. It can be observed that the planner with an incorporated occlusion tracker has already reduced its velocity to $1.5\text{m}\cdot\text{s}^{-1}$ to anticipate vehicles potentially coming from the right. The planner without occlusion tracking is still driving at its original velocity, not being aware of this oncoming danger. At $14$s, Fig.~\ref{fig:sim:scenario_0_overview}c shows that the planner with occlusion tracker has resumed driving after having observed the previously occluded vehicle and letting it pass by. At this point, the vehicle steers into the left turn, after which it will proceed its route behind the other vehicle. The planner without occlusion tracker, observed in Fig.~\ref{fig:sim:scenario_0_overview}e at $13.6$s, observes the occluded vehicle at $10$s, after which a slightly more severe braking action follows. The planner is capable of avoiding collision up to a $0.2$m margin, showing that even without occlusion planning, collision avoidance can be achieved. However, by the inclusion of occlusion tracking, the vehicle doesn't cross the intersection until it is clear that no other vehicle is coming, which from a risk-averse perspective is preferred. 

\begin{figure}
    \centering
    \includegraphics[width=1\linewidth]{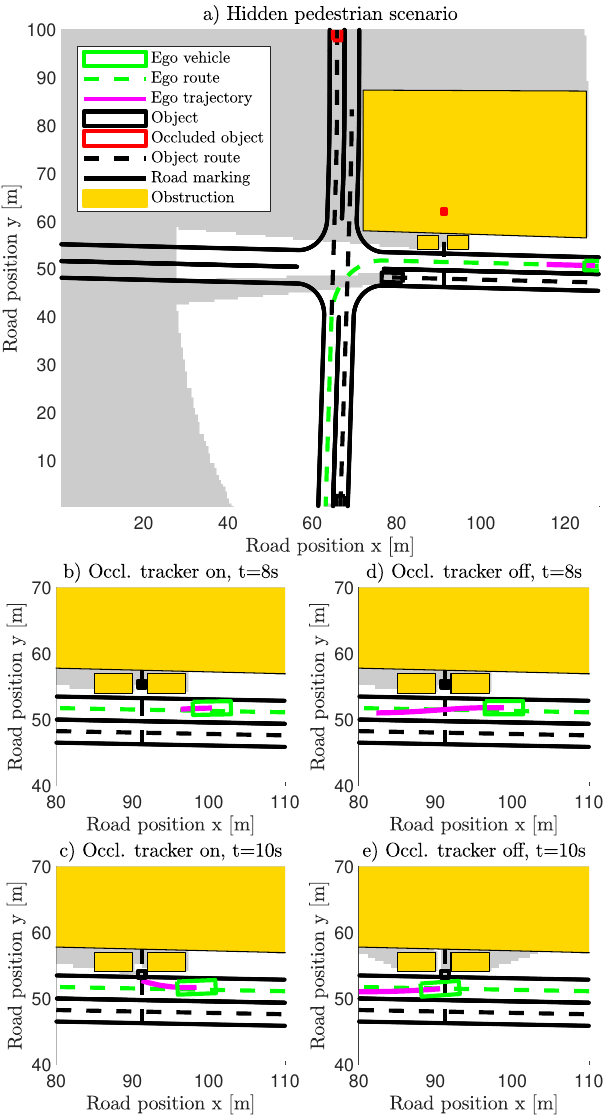}
    \caption{Overview of the "hidden pedestrian" scenario and snapshots at several time instances for motion planning results with and without occlusion tracking.}
    \label{fig:sim:scenario_1_overview}
\end{figure}
\begin{figure}
    \centering
    \includegraphics[width=1\linewidth]{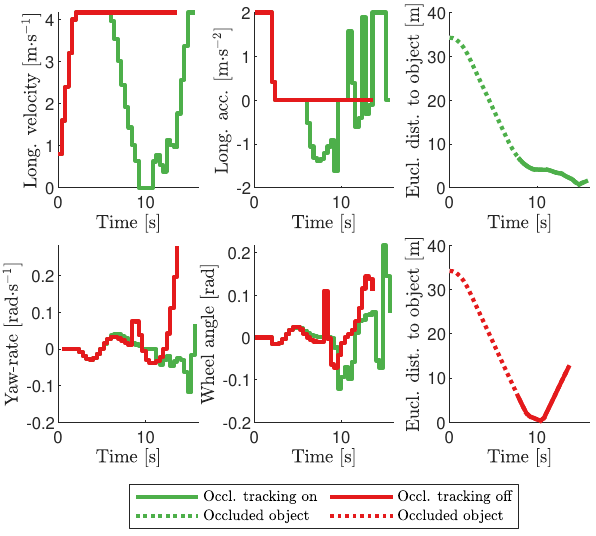}
    \caption{Signal plots for the "hidden pedestrian" scenario.}
    \label{fig:sim:scenario_1_plots}
\end{figure}
\subsection{Scenario 2: Hidden pedestrian}
The overall setup of the second scenario, where a hidden pedestrian appears from between two parked cars, is given in Fig.~\ref{fig:sim:scenario_1_overview}a. We consider two object classes, pedestrians and vehicles (i.e., $N_h=2$), where the considered maximum velocity of a hidden pedestrian is assumed to be~$v_p=0.84\text{m}\cdot\text{s}^{-1}$ and the maximum velocity of a vehicle is assumed to be $v_o=4.44\text{m}\cdot\text{s}^{-1}$. Pedestrians are assumed to spawn from pre-defined areas, in this case, the area between the two parked cars. Moreover, pedestrians are not assumed to be restricted in their direction of motion. Therefore, the structuring element $\set{D}^{a,h}$ (formed by overapproximating $\set{U}^{a,h}$~\eqref{eq:structuringelement}) used for pedestrians is a circular element with a radius of $v_p\cdot t_s$. Other than the ego vehicle, the simulation is populated by a vehicle and a pedestrian, where the pedestrian, walking at a speed of $v_p=0.84\text{m}\cdot\text{s}^{-1}$, is initially hidden from view. The vehicle, driving at a velocity of $v_o=4.44\text{m}\cdot\text{s}^{-1}$, is driving straight in the lane opposite of the driving lane for the ego vehicle. Until $6$s, the planners plan similar trajectories, driving straight while slightly avoiding the vehicle oncoming in the opposite lane as can be observed from Fig.~\ref{fig:sim:scenario_1_plots}. At $8$s, the time at which the hidden pedestrian is detected at a distance of $6.5$m, the occlusion-tracking planner (Fig.~\ref{fig:sim:scenario_1_overview}b) has already reduced its velocity with $65\%$ to anticipate potential oncoming pedestrians. The planner without occlusion tracking (Fig.~\ref{fig:sim:scenario_1_overview}d), detecting the pedestrian at a $5$m distance, is still driving at its initial velocity, however. From $9.2$s, the occlusion-tracking planner shortly comes to a full standstill to wait for the pedestrian to cross and assess whether no other objects are coming from the occluded zone. The planner without occlusion tracker makes an evasive steering maneuver to avoid the pedestrian and disregards any other potential hidden objects. At the time of $10$s, the same planner (Fig.~\ref{fig:sim:scenario_1_overview}e) successfully avoids the pedestrian, although with a very tight margin of $0.3$m, whereas the occlusion-tracking planner (Fig.~\ref{fig:sim:scenario_1_overview}c) allows the pedestrian to cross before proceeding with its route and overtakes the pedestrian from the right-hand side, showing a more preferable risk-averse strategy. 

\begin{figure}
    \centering
    \includegraphics[width=1\linewidth]{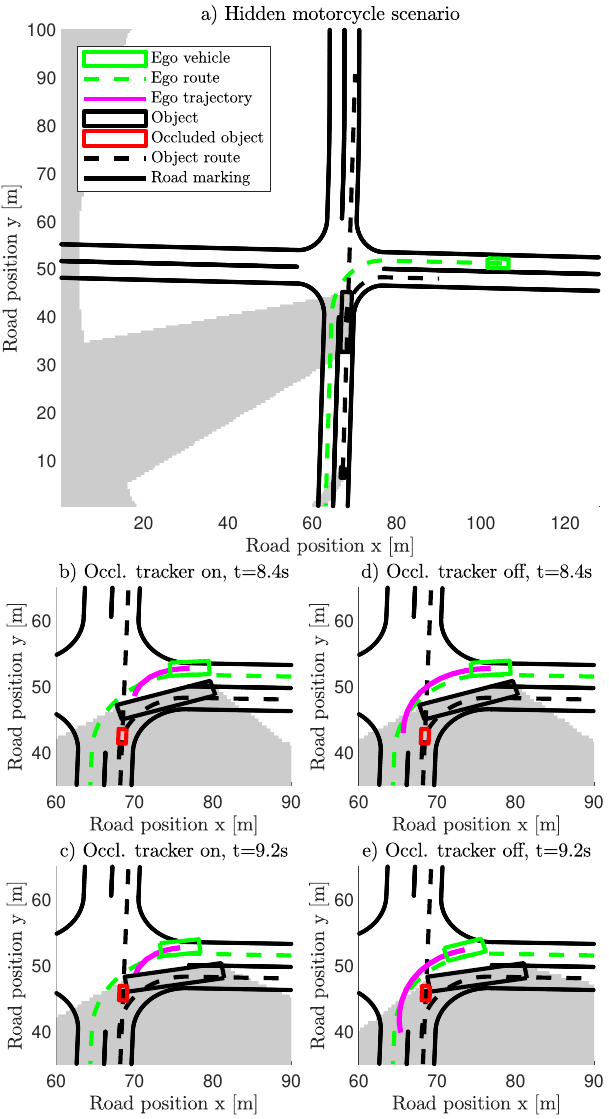}
    \caption{Overview of the "hidden motorcycle" scenario and snapshots at several time instances for motion planning results with and without occlusion tracking.}
    \label{fig:sim:scenario_2_overview}
\end{figure}
\begin{figure}
    \centering
    \includegraphics[width=1\linewidth]{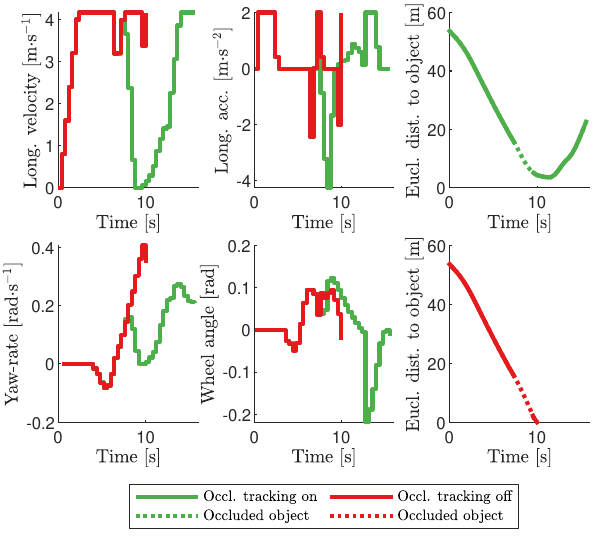}
    \caption{Signal plots for the "hidden motorcycle" scenario.}
    \label{fig:sim:scenario_2_plots}
\end{figure}
\subsection{Scenario 3: Hidden motorcycle}
The overall setup of the third and last scenario is given by the situation in Fig.~\ref{fig:sim:scenario_2_overview}a. Here, a crossing motorcycle approaches the intersection behind a truck. Once the truck takes a right turn, the motorcycle overtakes and gets visually occluded by the truck with respect to the ego-vehicle which is approaching the same intersection. We consider a single object class (i.e., $N_h=1$), where the considered class is that of vehicles (including motorcycles). It is assumed that these vehicles travel at a maximum velocity of $v_o=4.5\text{m}\cdot\text{s}^{-1}$. Until the time of $7.2$s, both planners show identical behavior, as can be observed in Fig.~\ref{fig:sim:scenario_2_plots}. That is, they accelerate from a standstill, travel across the lane, and close to the intersection brake for the truck, which is briefly entering their lane due to its dimensions. At a time of $8.4$s, it can be observed that the occlusion-tracking planner (Fig.~\ref{fig:sim:scenario_2_overview}b) has braked and has almost come to a full standstill to anticipate potentially oncoming hidden vehicles from behind the truck. The planner without occlusion tracking (Fig.~\ref{fig:sim:scenario_2_overview}d), however, has accelerated back to its original starting velocity and shows to be on a course of collision with the hidden motorcycle. Fig.~\ref{fig:sim:scenario_2_overview}c and Fig.~\ref{fig:sim:scenario_2_overview}e depict both planners, a single time step before the motorcycle is detected. For the occlusion-tracking planner, the vehicle has come to a full standstill to wait for any potentially hidden vehicles coming from behind the truck. The planner with occlusion planner however is still driving at its full velocity with only a small margin before the vehicles collide. At $9.6$s, both planners discover the presence of the motorcycle. However, the planner without occlusion tracking is only at a distance of $1.6$m from the motorcycle, as it attempts a brief braking action. Due to the small inter-vehicle distance, at time $10$s, the vehicle crashes into the motorcycle. The planner with occlusion tracking detects the motorcycle, while at a standstill, at a safe distance of $5.2$m, allowing the motorcycle to drive past and proceed its route. In this scenario, a collision could not be avoided without occlusion tracking, showing the serious hazard of occluded objects in these types of traffic scenarios and the true added benefit of our proposed approach.

In summary, the proposed risk-averse trajectory generator is able to cope with three different complex scenarios, by incorporating the description of the road, observations and predictions of obstacles, and anticipation of potentially hidden objects and predictions thereof. In contrast to the planner lacking occlusion tracking, our approach not only averts potential accidents but also maintains a comprehensible behavior, which reduces the perceived risk from a passenger's point of view. It accomplishes this by occasionally adopting a more cautious driving approach when the presence of objects along the intended path is uncertain. Furthermore, the planner avoids staying in one place for too long, since areas that were previously obstructed become less obstructed over time.

\section{Conclusion}~\label{sec:conclusion}
To increase safety in highly interactive on-road scenarios where object occlusions can occur, a model-predictive trajectory generator has been developed which contributes several aspects to the current state of the art. The model-predictive planner uses artificial potential fields (APFs), employed as risk fields, to maneuver its way through traffic situations including all relevant entities. In contrast to existing APF approaches the proposed planner allows the inclusion of seen objects, infrastructural elements, and reachable sets of potentially hidden objects of arbitrary shape and size. These entities are modeled in the form of a risk field and are methodologically constructed in this work using the newly introduced notion of a risk kernel: a parametric description of the risk surrounding a certain entity. Finally, the planner is made occlusion-aware through the use of sequential reasoning and hidden object modeling. These contributions allow us to model and tackle complex urban traffic scenarios as has been demonstrated in a simulation study.

Future work includes experimental validation of the proposed approach. Furthermore, it is of interest to extend the notion of risk fields toward stochastic scenarios, where the previously assumed measurements of objects and predictions thereof are no longer deterministic.
    \printbibliography

\end{document}